\documentclass[review]{elsarticle}

\usepackage{lineno,hyperref}
\modulolinenumbers[5]

\makeatletter
\def\ps@pprintTitle{%
 \let\@oddhead\@empty
 \let\@evenhead\@empty
 \def\@oddfoot{\centerline{\thepage}}%
 \let\@evenfoot\@oddfoot}
\makeatother
\usepackage{times}
\usepackage{epsfig}
\usepackage{graphicx}
\usepackage{amsmath}
\usepackage{amssymb}

\usepackage{subfigure}
\usepackage{anysize}
\usepackage{url}
\graphicspath{{Figures/}}
\usepackage[flushleft]{threeparttable}
\newcommand{\tabincell}[2]{
\begin{tabular}{@{}#1@{}}#2\end{tabular}}
\usepackage{color}
\usepackage{multirow}

\begin{document}
\begin{frontmatter}

\title{A Fully Trainable Network with RNN-based Pooling}

\author[uow]{Shuai Li\corref{mycorrespondingauthor}}
\ead{shuailichn@gmail.com}
\author[uow]{Wanqing Li}
\ead{wanqing@uow.edu.au}
\author[uow]{Chris Cook}
\ead{ccook@uow.edu.au}
\author[uestc]{Ce Zhu}
\ead{eczhu@uestc.edu.cn}
\author[uestc]{Yanbo Gao}
\ead{gyb\_chn@163.com}
\cortext[mycorrespondingauthor]{Corresponding author}
\address[uow]{School of Computing and Information Technology, University of Wollongong, NSW 2522, Australia}
\address[uestc]{School of Electronic Engineering, University of Electronic Science and Technology of China, Chengdu 611731, China}

%
%

\begin{abstract}
Pooling is an important component in convolutional neural networks (CNNs) for aggregating features and reducing computational burden. Compared with other components such as convolutional layers and fully connected layers which are completely learned from data, the pooling component is still handcrafted such as max pooling and average pooling. This paper proposes a learnable pooling function using recurrent neural networks (RNN) so that the pooling can be fully adapted to data and other components of the network, leading to an improved performance. Such a network with learnable pooling function is referred to as a fully trainable network (FTN). Experimental results have demonstrated that the proposed RNN-based pooling can well approximate the existing pooling functions and improve the performance of the network. Especially for small networks, the proposed FTN can improve the performance by seven percentage points in terms of error rate on the CIFAR-10 dataset compared with the traditional CNN.
\end{abstract}


\end{frontmatter}


\section{Introduction}
Convolutional neural networks (CNNs) have recently achieved the state-of-the-art performance in many image analysis tasks \cite{krizhevsky2012imagenet,simonyan2014very,szegedy2015rethinking, pavel2017object}. It has also been shown to be very effective in extracting features  for action recognition and other tasks involving temporal data \cite{karpathy2014large, simonyan2014two, donahue2015long, xin2016arch}. In most (if not all) of the existing CNN architectures such as the ``AlexNet''\cite{krizhevsky2012imagenet}, ``VGGNet''\cite{simonyan2014very} and ``InceptionNet''\cite{szegedy2015rethinking}, pooling is an important component for aggregating local features and reducing computational burden. In some networks \cite{springenberg2014striving}, convolution with strides (larger than $1$) is used to reduce the dimension of features to achieve a function similar to pooling. Noticeably, pooling is the only component in a typical CNN architecture (without considering normalization layers which are mostly for fast training and convergence) that is completely engineered with prior knowledge (such as max pooling and average pooling) instead of learning from data. Since the power of CNNs comes from their ability to adapt to the data through learning, the natural question to ask is ``Would pooling become the bottleneck of the network performance, and could pooling be learned in a similar way as other components from data?''. To answer this question, this paper proposes a learnable pooling function based on recurrent neural units. Together with the convolutional layers and fully connected layers in CNNs, such a learnable pooling leads to a fully trainable network (FTN). Compared with the traditional CNNs, it has the benefit of being fully adapted to data and task. Experimental results have demonstrated that FTN can improve the performance, especially on small networks.

The main contributions of this paper are summarized as follows.
\begin{itemize}
\item We propose a RNN based pooling method which is learnable and can be trained with data and task. Together with the CNNs, this leads to a fully trainable network (FTN), which can be trained end-to-end.
\item Since RNNs are universal approximators, the proposed RNN based pooling can be trained to be optimal for FTN. We have shown that one RNN neuron is able to approximate the existing average and max pooling with a high accuracy. Therefore, it can be easily used to replace the existing pooling operations in the existing models and be further fine-tuned, in addition to being trained end-to-end.
\item With the proposed RNN based pooling, FTN have achieved the state-of-art performance in the classification tasks. Moreover, it has been shown that for small networks, the proposed FTN can always improve the performance.
\end{itemize}

The rest of the paper is organized as follows. The related work is discussed in Section \ref{relatedwork}. Section \ref{fullTN} presents the proposed method and experimental results are shown in Section \ref{experiments}. Conclusion is drawn in Section \ref{conclusion}.

%

%

\section{Related work}
\label{relatedwork}
\subsection{Pooling}
\label{refpool}
Motivated from biology \cite{hubel1962receptive} where responses of simple cells are fed into complex cell through some pooling operations, the spatial pooling approach has been found very useful in many computer vision tasks. Up to now, the most commonly used pooling methods are still the max pooling and average pooling. The max pooling selects the most salient feature in a pooling region while the average pooling treats the features in a pooling region equally. However, the features in a local pooling region may be heterogeneous, leading to a loss of information on weak features through max pooling and loss of discriminative information through average pooling \cite{boureau2011ask}. It has been shown in the research that such pooling methods cannot achieve the optimal performance due to this information loss \cite{boureau2011ask}. A theoretical analysis of max pooling and average pooling for classification is provided in \cite{boureau2010theoretical} based on the i.i.d. Bernoulli distribution assumption for binary features and the exponential distribution for continuous features. It shows that the pooling cardinality and sparsity of the features affect its classification performance and the performance highly depends on the distribution of the features which is hard to estimate.

In addition to the max pooling and average pooling, there are some other pooling methods reported in the literature. In \cite{zhao2014protected}, a protected pooling method was proposed where a concave function is used to combine the features. The concave function is designed to protect weak codes in order to preserve details. In \cite{zeiler2013stochastic}, a stochastic pooling method was proposed where a multinomial distribution formed from the activation values is used. In this way, the location with large values are picked as output more frequently than others. Similarly in \cite{shi2016rank}, a rank based pooling was proposed to emphasize information with high rank over others. In \cite{rippel2015spectral}, spectral pooling was proposed which preforms dimensionality reduction by truncating the representation in the frequency domain. These methods are all heavily engineered to certain functions irrespectively of datasets, tasks and architectures of the networks. On the other hand, the convolution with sliding strides larger than one pixel \cite{springenberg2014striving} can be regarded as an extra convolutional layer with a pooling operation which selects the value of a fixed location. The learning pooling in \cite{sun2017learning} further simplifies the convolutional operation with independent linear operation on each channel.

There are also methods proposed to combine different pooling functions. In \cite{yu2014mixed}, a mixed pooling method was proposed where max pooling and average pooling are randomly selected with a stochastic procedure. Similarly, in \cite{lee2016generalizing}, the max pooling and average pooling are combined in a tree structure. In \cite{feng2011geometric} and \cite{gulcehre2014learned}, a geometric $l_p$-norm pooling was proposed to generalize the max pooling and average pooling, which can be represented as $(\sum_{i=1}^N |x_{I_i}|^p)^{1/p}$. When $p=1$, $l_p$-norm pooling reduces to average pooling and when $p=\infty$, $l_p$-norm pooling reduces to max pooling.

Pooling is a process that maps the $N \times N$ to $1$ where $N \times N$ represents the size of the local pooling region. The existing pooling methods certainly lose information in this mapping process. Compared to the other layers in CNNs such as the convolutional layers and the fully connected layers, pooling is likely to become the bottleneck for a CNN to reach optimal performance. Moreover, different pooling methods are used in different CNN architectures. Even in one CNN architecture such as the ``InceptionNet''\cite{szegedy2015rethinking}, different pooling methods are used. While it is difficult to select an appropriate pooling strategy for a better performance, it is also hard to explain how  and why one pooling strategy works better than others. Therefore, a flexible pooling function that can be learned from data for each pooling layer of a network architecture is highly desired.

\subsection{Recurrent neural networks}

Recurrent neural networks (RNNs) \cite{jordan1997serial} and their variants such as long short-term memory unit (LSTM) \cite{hochreiter1997long} and gated recurrent unit (GRU) \cite{cho2014learning} have been shown to be capable of aggregating sequential information and recognizing patterns in a sequence\cite{donahue2015long}. A common framework of employing RNNs is the encoder-decoder model \cite{srivastava2015unsupervised}. The encoder RNNs map an input sequence into a fixed length representation and then the decoder RNNs, based on the representation, are used to perform different tasks such as prediction. In addition to the temporal modelling, RNNs have also been used in spatial modelling for scene analysis \cite{byeon2015scene} and image generation \cite{theis2015generative}. In \cite{byeon2015scene}, a two-dimensional RNN network is used to model the spatial dependency in an image for scene labeling. In \cite{gregor2015draw}, a deep recurrent attention writer (DRAW) neural network was proposed for iterative generation of complex images where RNNs are used to process the image and iteratively provide attention regions for reading and writing.

Since the objective of pooling is to aggregate features of a local region, it is possible to consider RNN as a pooling function, especially considering that RNNs are universal approximators (Turing-Complete \cite{siegelmann1992computational}) and can be trained based on the data together with other parameters in the network. However, in practice, if the number of neurons required to approximate the pooling function is too large, it will make the training process inefficient and thus affect the performance of the whole neural network adversely. In this paper, we show that one RNN unit with modified activation functions is able to well approximate the existing max and average pooling methods. Therefore, better performance can be expected by incorporating the proposed RNN based pooling into CNNs.

\section{The proposed fully trainable network}
\label{fullTN}
\subsection{Overview}
\label{overview}
\begin{figure}[t]
	\centering
	\subfigure[]{
	\includegraphics[width=0.2\hsize]{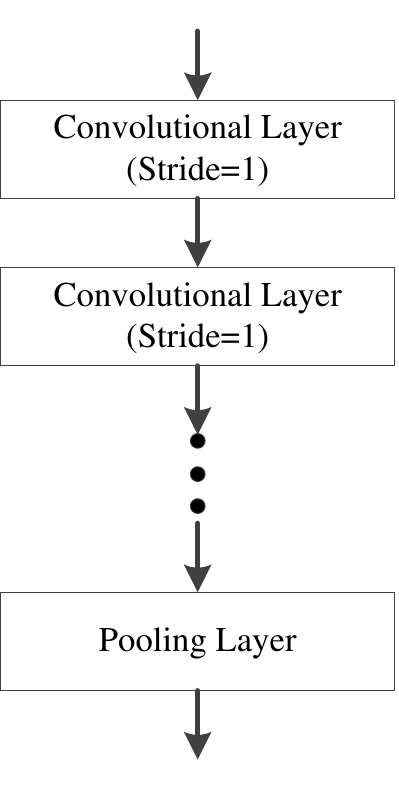}\label{components-pool}}\hspace{.8cm}
	\subfigure[]{
	\includegraphics[width=0.2\hsize]{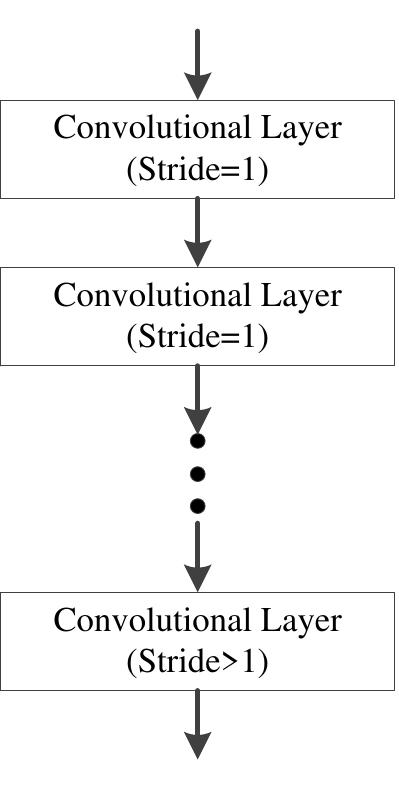}\label{components-conv}}\hspace{.8cm}
	\subfigure[]{
	\includegraphics[width=0.2\hsize]{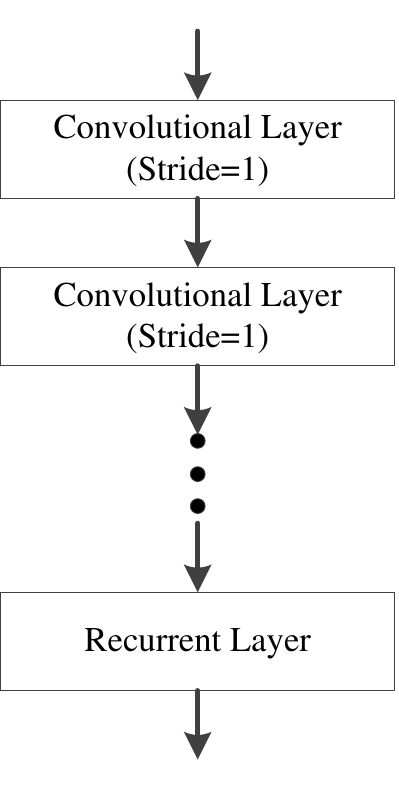}\label{components-rnn}}
	\caption{Key components in a CNN with (a) a pooling layer, (b) a convolutional layer with stride larger than 1, (c) a recurrent layer.}
\end{figure}

The basic component of a CNN is a stack of convolutional layers (usually more than 2) followed by a pooling layer as shown in Fig. \ref{components-pool}. The convolutional layer can be of many forms such as the traditional convolution structure \cite{simonyan2014very}, inception structure \cite{szegedy2015rethinking} and the residual structure \cite{he2016deep}. Normalization layers \cite{ioffe2015batch} may be used after or before each convolutional layer which is not considered here. The pooling layer is often a max pooling, average pooling or a pooling function as discussed above. Instead of using a pooling layer, a convolutional layer with stride larger than 1 can be used \cite{springenberg2014striving} as shown in Fig. \ref{components-conv} to reduce the dimension of the output features. In the proposed FTN, the basic component is a stack of convolutional layers followed by a recurrent layer as shown in Fig. \ref{components-rnn}. Specifically, the features in each pooling region are scanned into a sequence as input to the recurrent layer. There are many ways to perform the scan. The output of the recurrent layer at the last time stamp is the aggregated feature of the local pooling region, and so is treated as the pooled value. It has been empirically shown that the performance of the FNT is insensitive to the scanning order. Thus simple horizontal scanning is adopted in this paper. In addition to reducing the dimension of the features, the recurrent layer also intends to capture the pattern of the features in a local region. A FTN is constructed by stacking such components. 

Notice that in general any type of RNN can be used in the FTN for pooling. This paper adopts the commonly used LSTM unit. In the following, FTNs are explained in detail with respect to the extension of an LSTM unit for pooling, and the FTN architectures, respectively. 

\subsection{Extension of an LSTM unit for pooling}
\label{modifiedlstm}
In the study of CNNs, a general consensus is that for deep networks non-saturated activation functions such as rectified linear units (ReLU) are easier to train than the saturated activation functions such as logistic and hyperbolic tangent functions. In this paper, it is proposed to extend a conventional LSTM unit with non-saturated activation functions to preform pooling. Such an extension facilitates the training of the LSTM in a consistent way with other layers in a FTN.

\begin{figure}[t]
	\centering
  \includegraphics[width=0.6\hsize]{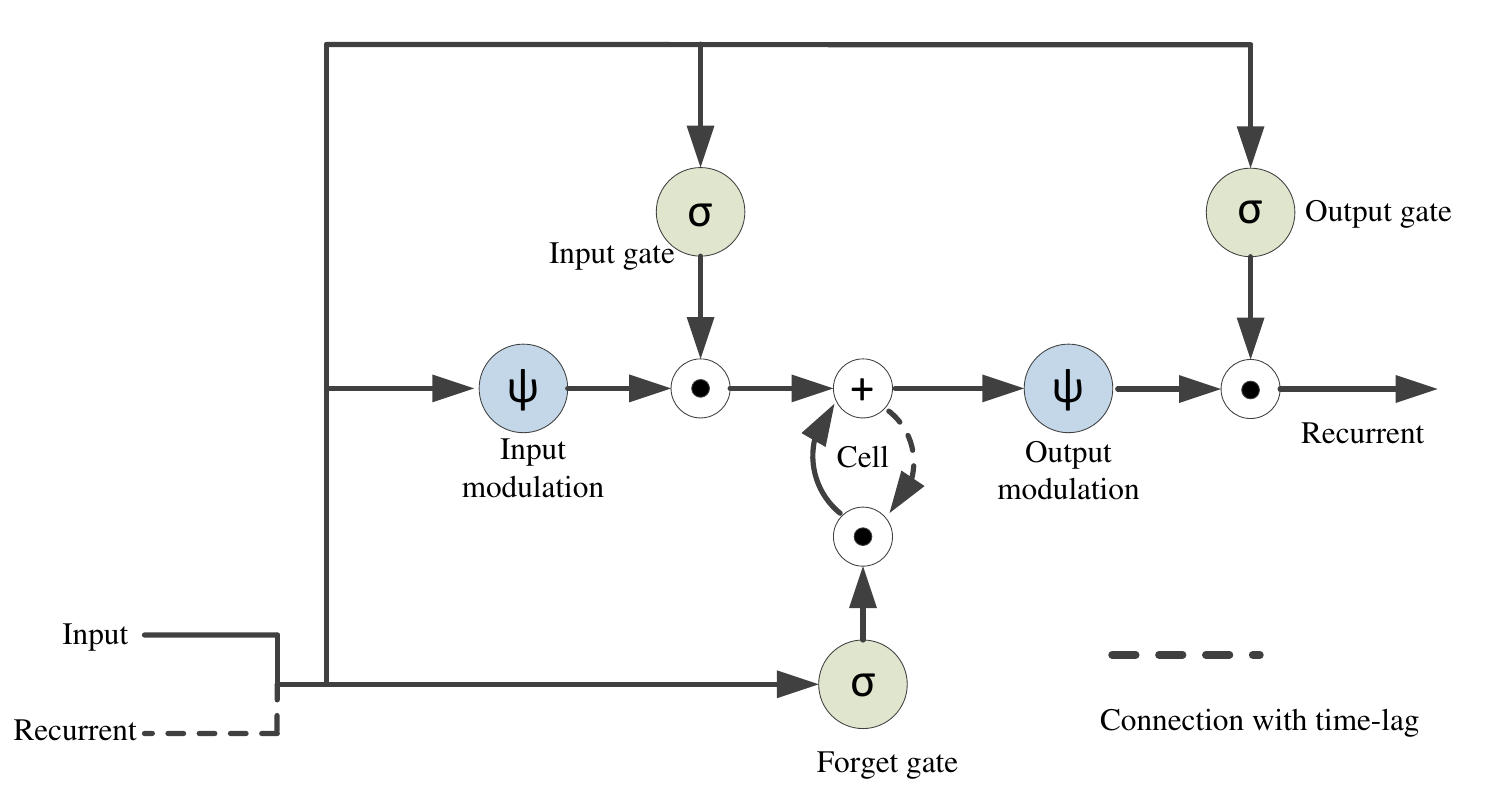}
	\caption{Illustration of a LSTM unit.}
  \label{lstm}
\end{figure}

The key component in LSTM \cite{hochreiter1997long} is a constant error carousel (CEC) which enforces a constant error flow over time steps. Fig. \ref{lstm} illustrates an LSTM without considering peephole connections. In addition to the CEC, LSTM contains three gates (input gate, forget gate and output gate), and two modulations (input modulation and output modulation). The gates are controlled by the current input and the recurrent input. The activation function for the gates are usually the sigmoid function ($\sigma$). The activation functions ($\psi$) used in input and output modulations are usually the hyperbolic tangent function ($tanh$). For input $\mathbf{x}$ at each time step $t$, the LSTM updates its states as follows: 

\begin{align}
    \mathbf{i}^{t} & =\sigma(\mathbf{W}_{\!i}\mathbf{x}^{t}+\mathbf{R}_{i}\mathbf{h}^{t-1} + \mathbf{b}_i) \nonumber\\
    \mathbf{f}^{t} & =\sigma(\mathbf{W}_{\!\!f}\;\!\!\mathbf{x}^{t}+\mathbf{R}_{f}\:\!\!\mathbf{h}^{t-1} + \mathbf{b}_{\!f\!}) \nonumber\\
    \mathbf{o}^{t} & =\sigma(\mathbf{W}_{o}\mathbf{x}^{t}+\mathbf{R}_{o}\mathbf{h}^{t-1} + \mathbf{b}_o) \nonumber\\
		\mathbf{g}^{t} & =\psi(\mathbf{W}_{\!g}\mathbf{x}^{t}+\mathbf{R}_{g}\mathbf{h}^{t-1} + \mathbf{b}_g) \nonumber\\		
		\mathbf{c}^{t} & =\mathbf{i}^{t}\odot\mathbf{g}^{t}+\mathbf{f}^{t}\odot\mathbf{c}^{t-1} \nonumber\\		
    \mathbf{h}^{t} & =\mathbf{o}^{t}\odot \psi(\mathbf{c}^{t})
    \label{lstmeq}
\end{align}
where $\mathbf{x}^{t}\in \mathbb{R}^{M}$, $\mathbf{h}^{t-1}\in \mathbb{R}^{N}$ and $M$, $N$ represent the dimension of the input feature at time step $t$ and the number of the neurons in LSTM, respectively. $\mathbf{i}^t$, $\mathbf{f}^t$ and $\mathbf{o}^t$ are the outputs of the input gate, forget gate and output gate, respectively. $\mathbf{g}^t$, $\mathbf{c}^t$ and $\mathbf{h}^t$ are the output of the input modulation, the cell state and the output of the LSTM, respectively. $\mathbf{W}_{v}$, $\mathbf{R}_{v}$ and $\mathbf{b}_{v}$ are the weight of the current input, the weight of the recurrent input, and the bias, respectively, for the input gate ($v=i$), forget gate ($v=f$), output gate ($v=o$) and input modulation ($v=g$). $\odot$ represents the point-wise multiplication.

With the hyperbolic tangent function used as the activation function for input and output modulations, the output of the LSTM is constrained to the range of $(-1, 1)$. However, the convolutional layers and fully connected layers in most CNN architectures employ non-saturated activation functions such as ReLU where their output ranges in $[0,+\infty)$. Therefore, the activation functions of the input and output modulations in a LSTM unit are required to be the same as those used in the convolutional layers.

However, change of the activation functions of the input and output modulations in a LSTM unit from a saturated function to a non-saturated function  would usually make the training of the LSTM hard to converge according to~\cite{le2015simple}. In this paper, one LSTM unit is applied to pool features in a local patch in each channel, where the dimension of input feature and the number of neurons for each channel are both $1$, i.e., $M=1$ and $N=1$. That is, the input, recurrent input, cell state and outputs of all the gates at each time instance and the corresponding weight parameters are all of dimension $1$. Let them be noted as $x^{t}$, $h^{t}$, ${i}^{t}$, ${f}^{t}$, ${o}^{t}$, ${c}^{t}$, $w_{\!g}$, $r_{\!g}$, $w_{\!v}$, $r_{\!v}$, $b_{\!v}$, respectively. The following proposition is used as a regulation in the training of LSTM.

\textbf{Proposition}: $w_{\!g}>0$ is a necessary condition for a LSTM neuron with ReLU activation function to converge when processing non-negative input features ($x^{t}$). 

{\it Proof:} (Proof by Contradiction) Let the bias of the input modulation be first ignored and considered later, that is, $g^{t} =\psi(w_{\!g}x^{t}+r_{g}h^{t-1})$, where $\psi$ is the ReLU activation function and $x^{t}\geq 0$. The initial state of the recurrent input ($h^0$) and cell ($c^0$) are both set to be $0$ which is used in most networks. Assume $w_{\!g}\leq 0$. Starting with time instance $1$, with $x^1\geq 0$, the output of the input modulation $g^1$ is zero. Since the outputs of all the gates including input gate, output gate and forget gate are non-negative, the output of LSTM ($h^1$) and the cell state ($c^1$) is $0$. Hence, the recurrent input and the cell state for the next time instance $2$ remains $0$. Together with $x^2\geq 0$, the output stays $0$. It can be deduced that under such circumstances, the output of LSTM stays $0$, which cannot be trained to converge. Therefore, the assumption does not hold and the opposite proposition ($w_{\!g}>0$) is true. On the other hand, bias determines the threshold to activate a neuron. For a LSTM unit, a negative bias for the input modulation deactivates the neuron for small inputs, making the neurons incapable of processing small features. Therefore, the bias for the input modulation is suggested to be constrained to be non-negative as well, in order to preserve the unit's ability of dealing with small inputs. 

Experimental results have shown that a LSTM unit with ReLU activation function can be trained robustly if this proposition is met.


\subsection{FTN architectures}
\label{ftnarch}
The proposed LSTM based pooling can be integrated with convolutional layers in different ways to create different FTN architectures. 
\begin{itemize}
\item One FTN architecture is that each local pooling region has its own LSTM to be trained and these LSTM units can work as different pooling functions for different regions. In this FTN, pooling is adaptive to local regions. 
\item The second FTN architecture has one LSTM per layer that is shared by all local regions in the layer. In this case, one pooling operation is performed on all local regions. However, pooling at different layers can be different depending on the training. For instance, the LSTM in one layer may act like a max pooling and the LSTM in another layer may act like an average pooling or a different function that the LSTM would best approximate for the training data.
\item The third FTN architecture is one LTSM unit shared by all pooling layers. This is equivalent to the conventional CNN where either max or average pooling is adopted. 
\end{itemize}
Obviously, the first architecture has the maximum number of parameters to be trained for the pooling and so is not considered further. In the experiments, the second and third FTN architectures are evaluated and compared to illustrate the benefits of the proposed LSTM based pooling over the traditional pooling. 

\section{Experimental Results}
\label{experiments}
Experiments were conducted to validate that a LTSM unit can well approximate max and average pooling functions and to verify the performance of the proposed FTN on tasks such as classification.

\subsection{LSTM for average and max pooling}
\label{lstmformaxexp}
The average pooling is a simple linear function which can be easily approximated by LSTM. However, the max pooling function is a highly non-linear function. In the following, an experiment was devised to show that one LSTM unit is able to approximate the max pooling function to a high degree of accuracy.

\textbf{Simulation setup}: ReLU was assumed as the activation function for the convolutional layers that a LSTM unit would work with. Experiments for other non-saturated activation functions can be conducted in a similar way. For ReLU ($max(0,x)$), the range of the output in theory is $[0,+\infty)$. In experiments of image classification, it is observed that the outputs generally fall in the range $[0, 300]$. Therefore, random numbers in this range were generated as the input to simulate the output of a convolutional layer. Since the pooling sizes used most in CNN are $2 \times 2$, $3 \times 3$ and $4 \times 4$, three sets of experiments were conducted with lengths of the input being $4$, $9$ and $16$, respectively. One LSTM unit with the modified activation function (ReLU here) was used.

\textbf{Training}: The LSTM was trained by minimizing the mean absolute error (MAE) between the output and the max value of the input to approximate the max pooling using mini-batch gradient descent with Nesterov momentum \cite{sutskever2013importance} and the batch size was set to 128. The initial learning rate was set to $0.1$ and the momentum was set to $0.9$. Regularizations such as weight decay and dropout were not used since infinite training examples can be generated. $10^4$ batches were considered as an epoch and one epoch was used for validation. The learning rate was decreased by a factor of $10$ when the validation accuracy stopped improving. The input weight and bias of the input modulation in the LSTM unit was initialized and regulated as described in subsection \ref{modifiedlstm}.

\textbf{Testing}: The batch size used for testing is the same as that for training. One epoch ($10^4$ batches) of data was generated for testing. The performance of the trained network is evaluated on three sets of input data: T1: random numbers in the range $[0, 300]$; T2: 50\% of random numbers in the range $[0, 300]$ and the other 50\% being $0$; T3: 20\% of random numbers in the range $[0, 300]$ and the other 80\% being $0$. The tests were designed to simulate the cases of general patches, relatively sparse patches and highly sparse patches considering that the responses of the convolutional neurons can be sparse. The performance for different pooling sizes are tabulated in Table \ref{lstm_maxpoolingresult}.

\begin{table}
\caption{MAE($10^{-5}$) of one LSTM unit with the modified activation function to approximate a max pooling function.}
\begin{center}
  \begin{tabular}{p{2.7cm} p{1.2cm} p{1.2cm}  p{1.2cm}}
  \hline
  & T1 & T2 & T3\\
  \hline
  Pool size $2 \times 2$ &$8.97$ & $8.91$ & $9.19$ \\
  \hline
  Pool size $3 \times 3$ &$4.42$ & $4.39$ & $4.40$ \\
  \hline
  Pool size $4 \times 4$ &$5.41$ & $5.28$ & $5.32$\\
  \hline
  \end{tabular}
\end{center}
\label{lstm_maxpoolingresult}
\end{table}

From Table \ref{lstm_maxpoolingresult}, it can be seen that one LSTM unit is able to well approximate the max pooling function as the errors are all smaller than $10^{-4}$ (which is negligible compared to the data range of $[0, 300]$). It can be also seen that the performance is insensitive to the pooling sizes.

\begin{table}
\caption{Classification result comparison on CIFAR-10 in terms of test error rate (\%) using different sizes of networks.}
\begin{center}
  \begin{tabular}{p{2cm}  c  c  c c}
  \hline
  Network & {max pooling} & {average pooling} & {proposed pooling (shared)} & {proposed pooling} \\
  \hline
  Conv\_8 & $57.44$ & $56.18$ & $50.96$ & $50.52$\\  
  \hline
  Conv\_16 & $32.75$ & $32.86$ & $28.58$ & $25.72$\\
  \hline
  Conv\_32 & $18.77$ & $20.82$ & $16.11$ & $15.48$\\
  \hline
  Conv\_64 & $13.27$ & $14.75$ & $12.04$ & $11.83$\\
  \hline
  \end{tabular}
\end{center}
\label{result_cifar10}
\end{table}

\subsection{Analysis of the LSTM based pooling}
\label{analysisRNNexp}
\subsubsection{Performance on different sizes of networks}
To illustrate the effectiveness of the proposed LSTM based pooling, experiments were conducted on the popular CIFAR-10 dataset. The dataset was preprocessed in the same way as in \cite{clevert2015fast}. That is, the dataset is preprocessed with global contrast normalization and ZCA whitening, and the images were padded with four zero pixels at borders. While training, $32 \times 32$ random crops with random horizontal flipping were used as input.

A CNN, composing of two stacks of $3 \times 3$ convolutional layers (2 layers in each stack) with a pooling layer at the end of each stack, 2 fully connected layers, and an additional fully connected layer of 10 units together with a softmax output layer for classification was used. The same number of units, denoted by $N$, were used in the convolutional layers and the $2$ fully connected layers. The corresponding network is denoted as Conv\_N.  To better demonstrate the effectiveness of the LSTM based pooling, a large local pooling region, namely $4 \times 4$ and $8 \times 8$ for the first and second pooling layers, respectively, were used. After the pooling layers, the size of the input to the fully connected layers is $1 \times 1$, thus the fully connected layers work in the same way as convolutional layers. A leaky ReLU unit with leakiness of $0.3$ was used as the activation function for both convolutional layers and fully connected layers, which has been reported \cite{mishkin2015all} to achieve a good performance on classification. Batch normalization \cite{ioffe2015batch} was used for convolutional layers and dropout (drooping rate 50\%) was applied after each fully connected layer. Total norm constraint on the gradients as in \cite{sutskever2014sequence} was used to stabilize the training. The initial learning rate was set to 0.01 and decreased by a factor of 10 after 50k and 40k iterations, respectively, and the training ended at the 122K-th iteration. SGD with Nesterov momentum \cite{sutskever2013importance} of 0.9 was used for training, and the batch size was $100$.


The results are shown in Table \ref{result_cifar10}. The column of ``proposed pooling'' and ``proposed pooling (shared)'' represent the second and third architectures described in subsection \ref{ftnarch}, respectively. That is to say, for ``proposed pooling (shared)'', both pooling layers share one LSTM unit while for ``proposed pooling'', each pooling layer has one LSTM unit. From the table, three observations can be made:

\begin{itemize}
\item The networks with the proposed LSTM pooling always improve the accuracy (lower the error rate) compared with the corresponding CNNs coupled with the traditional max pooling or average pooling function. 
\item When the network is very small such as Conv\_8, the performance improvement due to the LSTM based pooling is significant, up to $7$ percentage points. As the network size increases, the improvement decreases. It is conjectured that although a fixed pooling function may not optimally aggregate the local features, extra convolution kernels may compensate this. Thus with the increase of the convolution units, the gain of using a better pooling function over traditional pooling method drops.
\item The performance of the second architecture of FTN is better than the third, i.e., different LSTM units for different pooling layers improves the performance of FTN. This indicates that the optimal pooling functions for different pooling layers are likely to be different.
\end{itemize}


%

\begin{figure}[t]
	\centering
	\subfigure[]{
	\includegraphics[width=0.4\hsize]{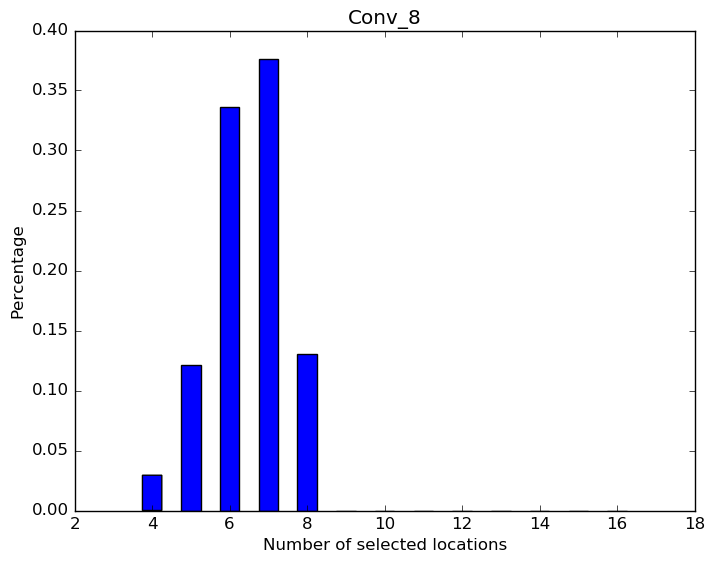}\label{max8}}
	\subfigure[]{
	\includegraphics[width=0.4\hsize]{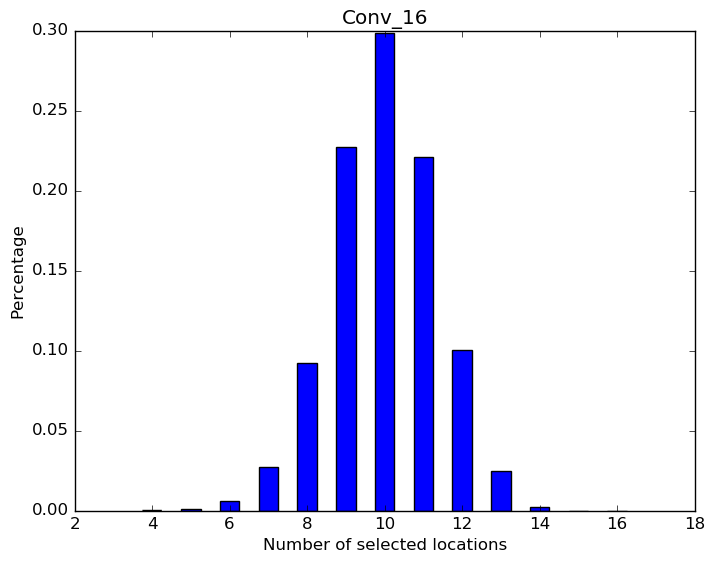}\label{max16}}
	\subfigure[]{
	\includegraphics[width=0.4\hsize]{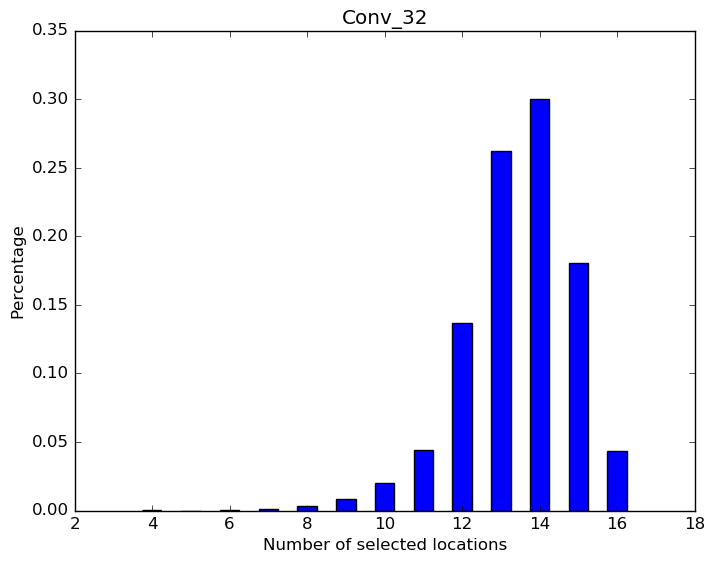}\label{max32}}
	\subfigure[]{
	\includegraphics[width=0.4\hsize]{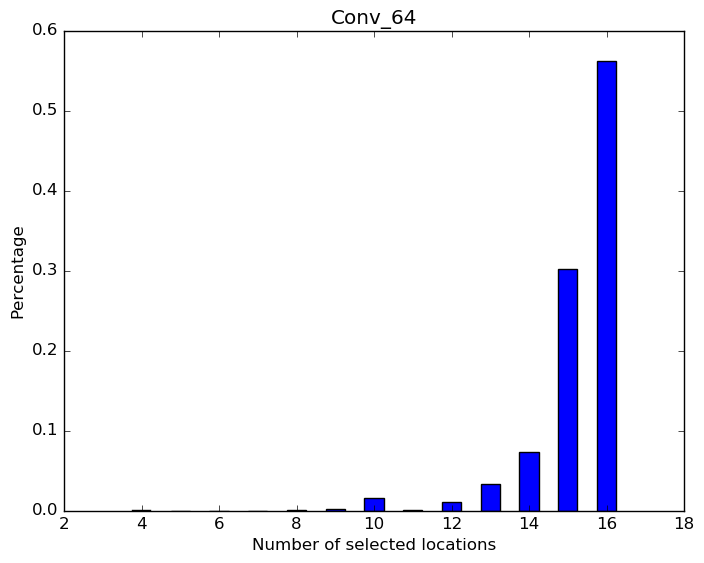}\label{max64}}
	\caption{Illustration of the location selection in the max pooling over different sizes of the networks.}
	\label{maxsel}
\end{figure}

%

\begin{figure}[htbp]
	\centering
	\subfigure[]{
	\includegraphics[width=0.48\hsize]{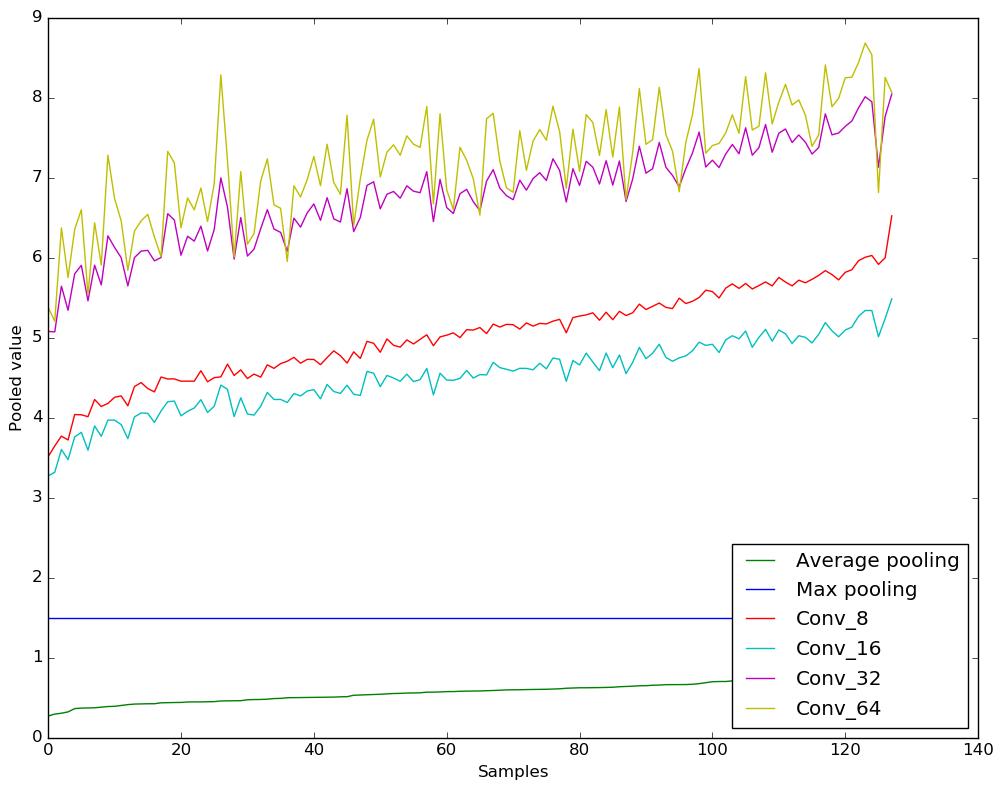}\label{l1_2}}
	\subfigure[]{
	\includegraphics[width=0.48\hsize]{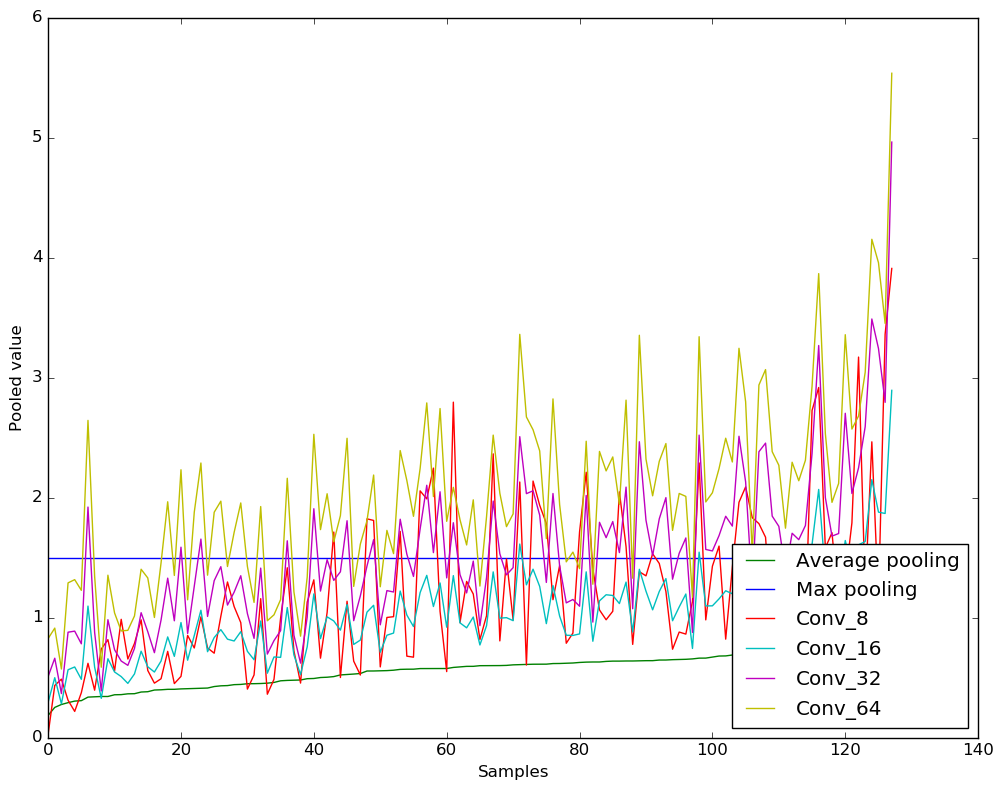}\label{l2_2}}
	\caption{Outputs of the learned pooling function from networks of different sizes in comparison with the max pooling and average pooling. (a) Learned function of the first pooling layer, (b) learned function of the second pooling layer.}
	\label{learnpool}
\end{figure}

\subsubsection{Analysis of the learned pooling function}
As shown in Table \ref{result_cifar10} and described above, the performance gap between the proposed pooling and the existing pooling methods becomes smaller with the increase of the number of convolution kernels, especially for the max pooling. Since pooling is a $N \times N$ to $1$ mapping process, information of certain locations in the pooling region may be lost in the existing pooling process, leading to a degraded performance of the network. In the following, we first show that networks are trained to preserve information of different locations in a pooling region.

For a CNN, max pooling is used independently for each channel. And for each channel, it selects the value of one location (which is the location with the maximal value) as the output and the information at that location is implicitly carried forward in this channel. That is to say, for the pooling process over multiple channels, information from a number of locations may be selected and preserved by one or more channels. Fig. \ref{maxsel} shows the histogram of the number of locations that have been selected by at least one channel. The output of max pooling for 5000 randomly selected local patches in the dataset are used for different networks. The pooling size is $4*4$ in all networks, leading to a local region of 16 locations. For the Conv\_8 network, the number of neurons is 8, and thus at most 8 locations can be selected by the max pooling (when locations selected by different channels are all different). Fig. \ref{max8} shows that generally more than 4 locations have been selected in one or more than one channels, and for some pooling regions, all 8 locations are selected. For the Conv\_16 network shown in \ref{max16}, generally more than 6 locations have been selected. For the Conv\_64 network shown in \ref{max16}, in over 50\% pooling regions, all 16 locations have been selected. It is reasonable to assume that when the number of the neurons is large enough, information from all locations may be implicitly carried forward after the pooling operation. It indicates that networks (convolution kernels) are trained to sample information from all locations. Consequently, with the increase of the number of neurons, the effect of a good pooling operation may be reduced since information from more locations could be sampled through different channels. However, in this case, more training data are required to train the increased number of parameters. On the other hand, since LSTM can aggregate information of a sequence, it can adaptively sample more information from all the pooling locations than the existing pooling functions. Therefore, the performance of the proposed FTN is significantly better than the traditional CNN when the number of neurons used is small.

To illustrate the learned LSTM based pooling functions for different pooling layers in the network of different sizes, the output of the pooling function in comparison with the max pooling and average pooling is shown in Fig. \ref{learnpool}. For better illustration, random values with a fixed maximum value ($1.5$) is used as input and the output is rearranged according to the magnitude of the average pooling result. Fig. \ref{l1_2} shows the output of the learned pooling function from the first pooling layer. ``Conv\_N'' represents the outputs obtained from the different pooling functions learned from their corresponding ``Conv\_N'' networks, respectively. Note that the mean value of each pooling result can be compensated by the bias of the neurons in the following layer, thus the variation of each pooling result is more meaningful. It can be seen that the learned pooling functions of the first pooling layer work similarly as the average pooling. Especially for the small networks such as ``Conv\_8'' and ``Conv\_16'', the output highly correlates with the output of average pooling and the variation is relatively small. This indicates that average pooling may perform better than max pooling for the first pooling layer of small networks, which agrees with our results shown in Table \ref{result_cifar10}. 

For the learned pooling function of the second pooling layer as shown in Fig. \ref{l2_2}, it can be seen that the variation is very large, i.e., highly sensitive to the different patterns of the inputs. First, compared with the input to the first pooling layer, each input to the second pooling layer corresponds to a larger region of the original image and thus more useful information for the task. Second, it is known that outputs of the higher layers in the network capture high level information, and information at different locations may produce different contexts for the final classification task. For example, the same input with different orders may produce different results. Thus it is very important for the pooling layer to aggregate information while capturing useful patterns. This can be done using our proposed pooling while not possible for the traditional max pooling and average pooling. 

By comparing the learned pooling functions of two layers, it can be seen that the optimal pooling functions for different pooling layers are quite different. In the traditional CNNs, max pooling and average pooling are often selected empirically. In the ``InceptionNet'' \cite{szegedy2015rethinking}, both max pooling and averaging pooling are adopted at different layers also empirically. In such a case, the whole network cannot achieve the best performance. On the other hand, our proposed LSTM based pooling is able to be adaptive for each layer to the training data and thus achieve a better performance.

\begin{table}
\caption{Complexity comparison between CNN and the proposed FTN in terms of time (sec per batch).}
\begin{center}
  \begin{tabular}{p{2cm}  c  c  c c}
  \hline
   & \tabincell{c}{Train \\with cudnn} & \tabincell{c}{Train \\without cudnn} & \tabincell{c}{Test \\with cudnn} & \tabincell{c}{Test \\without cudnn} \\
  \hline
  CNN & $0.013$ & $0.021$ & $0.0027$ & $0.0037$\\  
  \hline
  FTN & $0.033$ & $0.035$ & $0.0068$ & $0.0076$\\
  \hline
  \end{tabular}
\end{center}
\label{time}
\end{table}

\subsubsection{Complexity} 
As mentioned in subsection \ref{overview}, the proposed LSTM based pooling first transforms the $N \times N$ local region to a sequential input of length $N \times N$. Then it processes this sequential input. For example, for the general $2\times 2$ pooling, LSTM needs to process inputs of 4 time steps. It is known that the update of LSTM at each time step in Eq. (\ref{lstmeq}) can be regarded as convolution with multiple channels. So the complexity of the proposed pooling is similar to performing 4 convolutional layers. To evaluate its complexity, a similar network as the Conv\_64 network (except that the pooling size is set to be $2\times 2$) was used. Batch size was set to be 1 to purely monitor the computation without considering memory issues. The program was implemented based on Theano \cite{team2016theano} and Lasagne, and runs on a TITAN X GPU. The time used in the training and testing process is shown in Table \ref{time}. Since convolution is heavily optimized in cudnn (the deep neural network library developed in NVIDA CUDA), we show both results obtained with cudnn and without cudnn. It can be seen that the complexity of training the above FTN network is about two-three times of training CNN. This is consistent with our above analysis that the complexity of the proposed LSTM based pooling is similar to performing 4 convolutional layers. Since pooling is only applied a few times depending on the size of the input (around 5 times for input of size 256), the complexity of training FTN is bounded. Compared to the current networks over 100 layers, the increase in training time is acceptable considering the benefit of having a learnable pooling function to improve the performance. Moreover, compared to the image modelling methods that use RNN to process the whole image in a sequential manner, the time increase due to the proposed pooling is relatively very small. It is worth noting that since the proposed pooling is learned from data for a network, it can be used as a tool to develop new pooling functions for different applications.


\begin{table}[t]
\centering
\caption{Comparison of the proposed FTN and CNNs on CIFAR-10 and CIFAR-100 in terms of test error rate (\%).}
\begin{threeparttable}
  \begin{tabular}{p{5cm} p{3cm}  p{2.5cm}  }
  \hline
  Network & CIFAR-10 & CIFAR-100 \\
  \hline
  DSN\cite{lee2015deeply} & $7.97$ & $34.57$ \\
  \hline
  NIN\cite{lin2013network} & $8.81$ & $35.68$ \\
  \hline
  Maxout\cite{goodfellow2013maxout} & $9.38$ & $38.57$ \\
  \hline
  All-CNN\cite{springenberg2014striving} & $7.25$ & $33.71$ \\
  \hline
  Highway Network\cite{srivastava2015training} & $7.60$ & $32.24$ \\
  \hline
  ELU\cite{clevert2015fast} & $6.55$ & $24.28$ \\
  \hline
  LSUV\cite{mishkin2015all} & $6.06$ & $29.96$ \\
  \hline
  LSUV\textsuperscript{*}\cite{mishkin2015all} & $5.84$ & N/A \\  
  \hline  
  LEAP\cite{sun2017learning} & $7.17$ & $29.80$ \\  
  \hline  
  Stochastic Pooling\cite{zeiler2013stochastic} & $15.13$ & $42.51$ \\
  \hline
  Rank based Pooling\cite{shi2016rank} & $13.84$ & $43.91$ \\
  \hline
  Mixed Pooling\cite{yu2014mixed} & $10.80$ & $38.07$ \\
  \hline
  Tree Pooling\cite{lee2016generalizing} & $6.67$ & $33.13$ \\
  \hline
  Tree+Max-Avg Pooling\cite{lee2016generalizing} & $6.05$ & $32.37$ \\
  \hline
  \textbf{Proposed FTN} & $5.79$ & $26.89$ \\
  \hline
  \multicolumn{3}{c}{With extreme data augmentation\cite{graham2014fractional}}\\
  \hline
  Fract. Max-pooling \cite{graham2014fractional} & $4.50$ & $26.39$ \\
  \hline
  All-CNN\cite{springenberg2014striving} & $4.41$ & N/A \\
  \hline
  \end{tabular}
    \begin{tablenotes}
      \small
      \item LSUV\textsuperscript{*} is obtained with deep residual network using maxout as activation function.
      \item N/A represents the result is not provided in the corresponding paper.
    \end{tablenotes}
\label{result_final}
\end{threeparttable}
\end{table}

\begin{figure}[t]
	\centering
	\subfigure[]{
	\includegraphics[width=0.48\hsize]{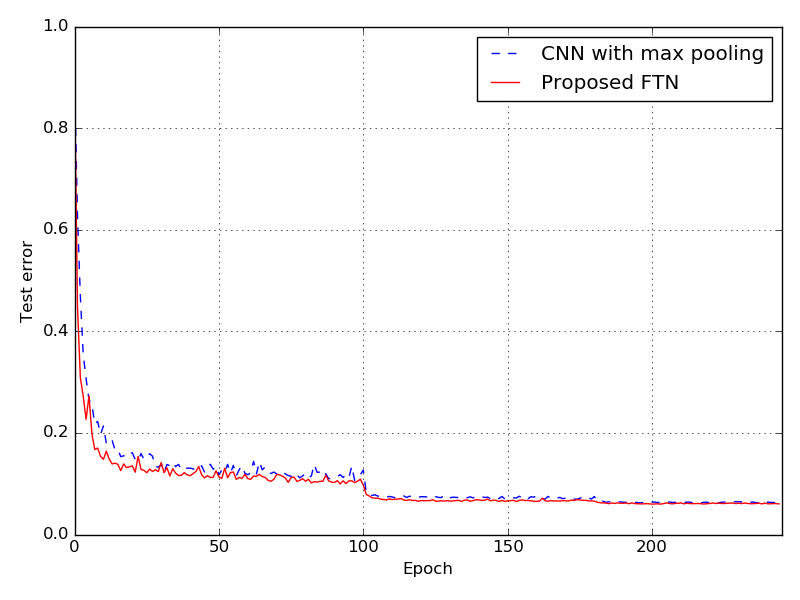}\label{testerrorfull}}
	\subfigure[]{
	\includegraphics[width=0.48\hsize]{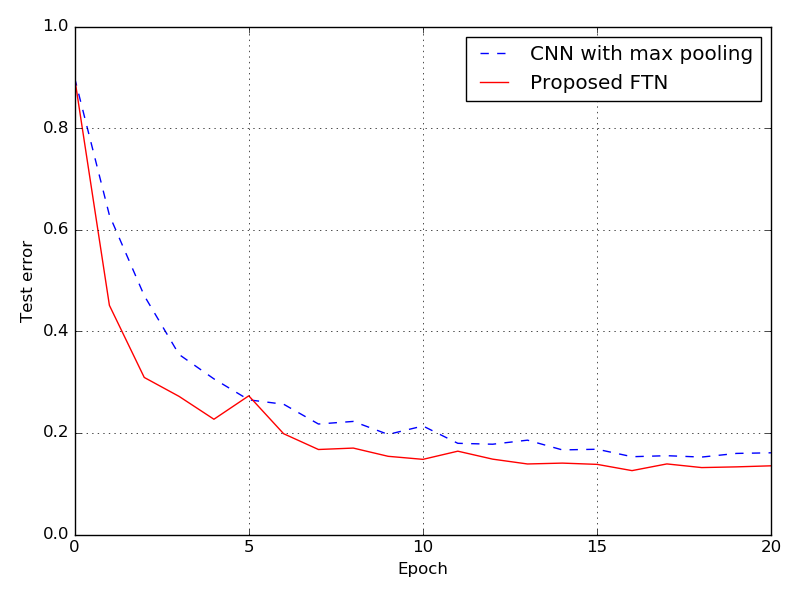}\label{testerrorpart}}
  \caption{Test error comparison between the proposed FTN and the traditional CNN, (a) over the entire training process, (b) over the first 20 epochs.}
\end{figure}

\subsection{Classification performance on CIFAR-10 and CIFAR-100}
The widely used CIFAR-10 and CIFAR-100 datasets were used to evaluate the performance of the proposed FTN (the second architecture). The VGG16 architecture \cite{simonyan2014very} was used for classification due to its popularity. It is composed of 5 stacks of convolutional layers with a $2 \times 2$ pooling layer at the end of each stack, and 2 layers of fully connected layers in the end. A fixed 10/100 units fully connected layer together with a softmax output layer are added for classification of CIFAR-10 and CIFAR-100, respectively. The leaky ReLU unit with leakiness of 0.1 was used as activation functions for both the convolutional layers and fully connected layers. Dropout was used after the pooling layers (dropping rate 30\%) and fully connected layers (dropping rate 50\%) to regularize the training. The preprocessing of the dataset and the training procedure are the same as in Subsection \ref{analysisRNNexp}. For the proposed FTN, the pooling layers were replaced with LSTM units, one for each pooling layer. It is worth noting that one LSTM unit only introduces $12$ parameters. On the contrary, one convolutional unit generally introduces $N \times N \times C_{l-1}+1$ units where $N$ is the kernel size, $C_{l-1}$ is the number of channels of the input to the current unit, and $+1$ indicates the bias. Compared to the large amount of parameters used in CNN, the increased number of parameters due to a LSTM unit is negligible.

The results of the FTN and the comparison to the state-of-the-art methods are shown in Table \ref{result_final}. It can be seen that under similar training conditions (without the extreme data augmentation \cite{graham2014fractional}), the proposed FTN achieves the state-of-the-art performance. 
Although LTSM has been reported to be difficult to train in the literature, it is found that the proposed FTN converges very fast in the experiments, even faster than a CNN with traditional max pooling. The test errors of the proposed FTN and CNN v.s. the training epochs are shown in Fig. \ref{testerrorfull}. Fig. \ref{testerrorpart} shows the zoomed-in curve of the testing errors of the first 20 epochs. It can be clearly seen that the proposed FTN achieved a relatively higher performance in less iterations than the CNN.

\section{Conclusion}
\label{conclusion}
In this paper, a fully trainable network (FTN) is proposed. Compared with the traditional CNNs, the handcrafted pooling layer is replaced with a LSTM unit in the proposed FTN. Due to the capability of a LTSM or RNN in general in modelling sequential data, the proposed learnable pooling can be trained to capture patterns of the data in the pooling regions. Specifically, we have shown that LSTM based pooling can approximate the existing pooling functions with a very high accuracy. Moreover, the proposed FTN can significantly outperform small traditional CNNs and achieve comparable performance to large CNNs.

\section*{References}
\bibliography{Reference}
\end{document}